# MGHF: Multi-Granular High-Frequency Perceptual Loss for Image Super-Resolution

Shoaib Meraj Sami[1], Md Mahedi Hasan[1], Mohammad Saeed Ebrahimi Saadabadi[1], Jeremy Dawson[1], Nasser Nasrabadi[2], Raghuveer Rao[3]
[1]West Virginia University, WV, USA [2]Johns Hopkins University, MD, USA
[3]DEVCOM Army Research Laboratory

## Abstract

While different variants of perceptual losses have been employed in super-resolution literature to synthesize more realistic, appealing, and detailed high-resolution images, most are convolutional neural networks-based, causing information loss during guidance and often relying on complicated architectures and training procedures. We propose an invertible neural network (INN)-based naive **M**ulti-**G**ranular **H**igh-**F**requency (MGHF-n) perceptual loss trained on ImageNet to overcome these issues. Furthermore, we develop a comprehensive framework (MGHF-c) with several constraints to preserve, prioritize, and regularize information across multiple perspectives: texture and style preservation, content preservation, regional detail preservation, and joint content-style regularization. Information is prioritized through adaptive entropy-based pruning and reweighting of INN features. We utilize Gram matrix loss for style preservation and mean-squared error loss for content preservation. Additionally, we propose content-style consistency through correlation loss to regulate unnecessary texture generation while preserving content information. Since small image regions may contain intricate details, we employ modulated PatchNCE in the INN features as a local information preservation objective. Extensive experiments on various super-resolution algorithms, including GAN- and diffusion-based methods, demonstrate that our MGHF framework significantly improves performance. After the review process, our code will be released in the public repository.

## 1 Introduction

Super-resolution (SR) aims to improve the detailed information in images degraded by downsampling, blurring, noise, and various real-world distortions [1; 2]. Degraded images contain structural information but lack high-frequency information [3; 4]. Researchers employ various generative models [2; 5; 6; 7; 8; 9] and objective functions [10; 11; 12; 13; 14] to enhance high-frequency features in the SR problem [2; 6; 15]. The objective functions for SR can be categorized as perceptual [10; 11], content [16], style losses [17], structural similarity measures [18; 19], and frequency domain losses [20; 21]. Among these categories, naive perceptual losses [10; 11] are widely used; however, while effective in capturing many characteristics of the source image, they fall short of preserving complete details due to the inherent information approximation [22; 23] and lossy nature of CNN operations [24]. In the SR literature [11; 14; 25; 26], several variants of information approximation within the perceptual loss family have been implemented through diverse techniques such as quantization [27], adversarial training [28], neural network feature extraction [22; 23; 29; 30], and feature enhancement [31]. Some of these approximation approaches are: i) LPIPS [11], which employs learned feature map weighting to align with human perception; ii) FDPL [26], which applies quantization to discrete cosine transform (DCT) [32] coefficients, despite DCT's inherent lossless nature; iii) Fourier space loss [25], which shifts generation toward perceptually pleasing high-frequencies through adversarial training [33]; and iv) wavelet domain style transfer [14], which introduces feature enhancement through a selective wavelet filter. Moreover, OSEDiff [2] inherently utilizes the LPIPS [11] objective that is based on CNN instead of a bijective mapping and lossless invertible neural network (INN) [34], which causes loss of information and approximation errors when calculating the perceptual objective. This phenomenon is revealed in our experimental results in Figure 6, where the INN-based MGHF framework improves OSEDiff performance across different metrics.

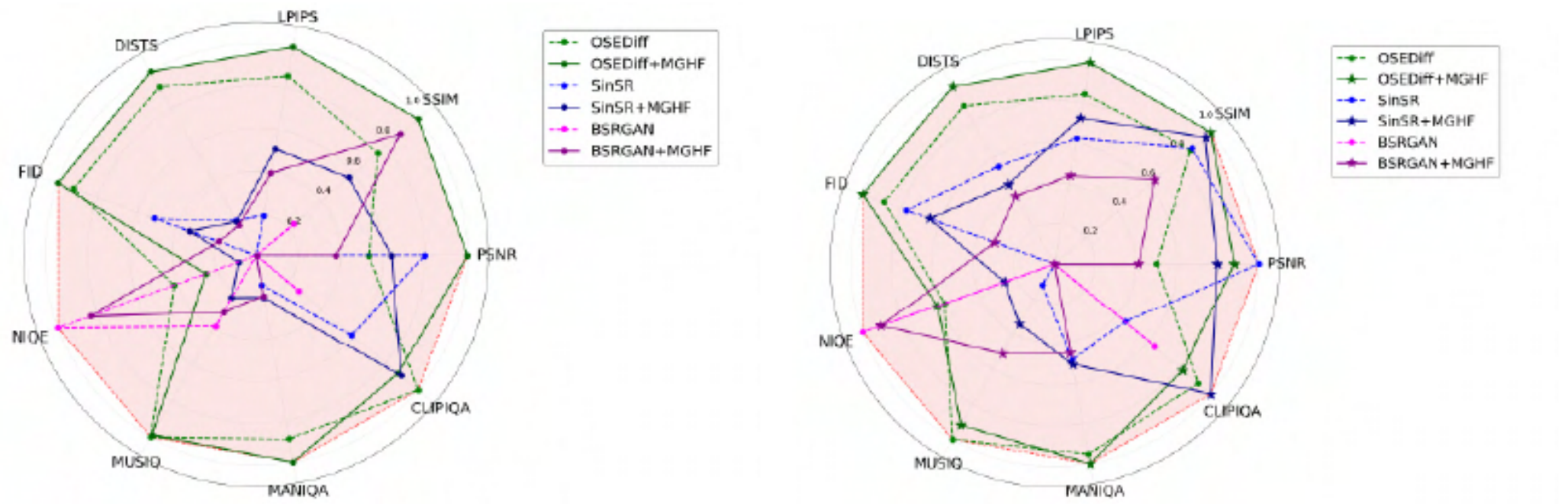

(a) Performance comparison on DrealSR (b) Performance comparison on RealSR

Figure 1: Performance comparison of different super-resolution models with and without MGHF framework. (a) Results on the DrealSR [37] dataset showing the effectiveness of MGHF across different metrics. (b) Results on the RealSR [38] dataset demonstrate consistent improvements.The dotted line of each color represents the baseline model, and the solid line of the same color represents the baseline model with the MGHF framework.

Another inherent problem of several perceptual loss families during SR is the substantial complexity of the architecture design [13; 35] and training procedure [5]. SRGAN [5] employs a relatively straightforward perceptual loss [10] using VGG [36] features, but requires unstable adversarial training [33]. SROBB [35] significantly increases complexity by introducing region-specific perceptual losses that process objects, backgrounds, and boundaries differently, requiring additional segmentation labels and specialized loss calculations for each semantic region. SR4IR [13] presents the complex training methodology with its alternate training framework that switches between updating the SR network and the task network, combined with a specialized cross-quality patch mix data augmentation strategy. We propose a naive multi-granular high-frequency perceptual loss that maintains an efficient architecture while delivering effective results for the super-resolution task to address these complexity issues.

Perceptual losses [10; 11] trained on the VGG [36] or AlexNet [39] backbone in ImageNet [40] and stable diffusion [41] trained on billions of image-text pairs serve as important super-resolution priors [2; 10; 11; 42]. We introduce a novel high-frequency perceptual loss based on an invertible neural network (INN) trained on ImageNet [40] as a new prior for SR. INNs have previously been utilized in image super-resolution and rescaling [43] literature in ways distinct from our approach. For instance, SRFlow [6] employs INN-based normalizing flows [44] to learn conditional distributions of high-resolution images given low-resolution inputs, while IRN [43] explicitly models downscaling or upscaling as forward or inverse operations of an invertible network with Haar wavelet [45] transformation. HCFlow [46] creates bijective mappings between HR-LR image pairs where high-frequency components are hierarchically conditional on low-frequency components through specially designed flow levels, and IARN [47] adapts the invertible framework by replacing Haar wavelet transforms with preemptive channel splitting and embedding position-aware scale encoding, enabling arbitrary rescaling factors within a single model while maintaining bidirectional invertibility. The authors [48] have introduced invertible priors for image rescaling through invertible feature recovery modules (IFRM), which establish bijective transformations between quantized features by VQGAN [49] and low-resolution images using coupling layers [34]. Our contribution is novel as we leverage an INN trained on ImageNet as an SR prior.

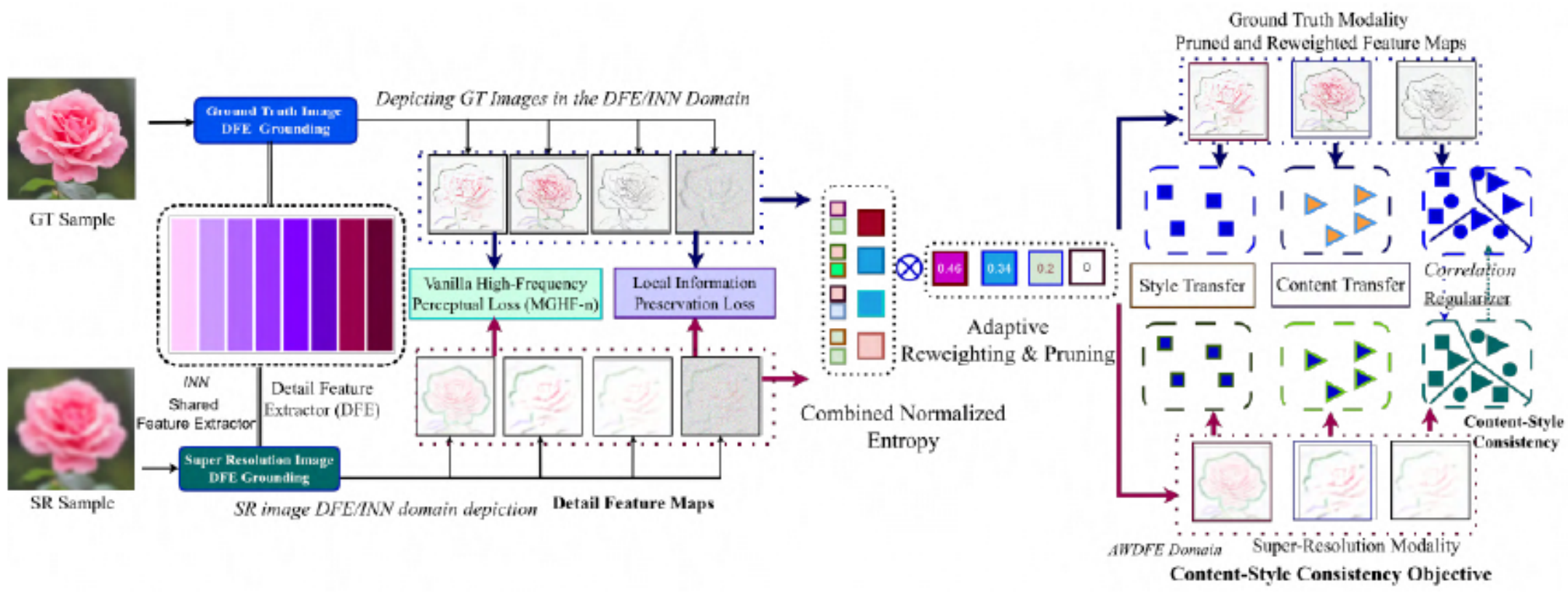

Figure 2: The depiction of proposed MGHF perceptual loss, where the detail feature extractor (DFE) is based on an invertible neural network. Vanilla high-frequency perceptual loss is calculated among feature maps of the DFE, while the content-style consistency loss is calculated from the most informative pruned and reweighted DFE feature maps.

We propose a **m**ulti-**g**ranular **h**igh-**f**requency perceptual loss (MGHF) to overcome the aforementioned issues. The naive version, MGHF-n, serves as an effective invertible neural network (INN) prior trained on ImageNet to guide the super-resolution process. Building upon this foundation, our comprehensive version (MGHF-c) addresses the perception-distortion tradeoff [50] and improves the SR performance on several image quality metrics [51; 52; 53] by both focusing and regularizing essential detail information alongside the INN prior. To achieve these goal, MGHF-c introduces an adaptive importance score based on normalizing entropy to prioritize and select significant INN features, which are then processed through a multifaceted approach that incorporates a modulated PatchNCE [54]-based local information preservation objective to maintain intricate details, while simultaneously preserving style and content information in the INN domain via Gram matrix and mean-squared loss, respectively. Additionally, to overcome unnecessary style transfer and preserve content information while guiding SR, we propose a correlation loss-based content-style consistency regularizer. Our experiments demonstrate that the proposed MGHF objective significantly improves the performance of three super-resolution algorithms: OSEDiff [2], SinSR [55], and BSRGAN [56], with the first two based on diffusion models and the last on a GAN. Notably, in SinSR [55], even our simpler variant, MGHF-n, outperforms both LPIPS [11] and the naive perceptual loss [57]. Furthermore, our proposed INN feature extractor within the MGHF framework requires 42 times fewer parameters than the VGG [36]-based feature extractor typically used for calculating perceptual losses [10; 11].

## 2 Related Works

### 2.1 Image Super-Resolution

Super-resolution is a well-known low-level computer vision problem widely used in many applications [1; 58], such as surveillance [59], medical imaging [60], gaming [61], virtual reality [62], photography [63], face recognition [64], etc. After the evolution of AlexNet [39], researchers implemented deep learning-based super-resolution approaches [57; 58]. Following that, the generative adversarial network (GAN) evolved, and the GAN-based SR algorithms [5; 56; 65] were mainstream in the computer vision community [58]. The SR-GAN [5], ESRGAN [65], and RankSRGAN [66] are some well-known GAN-based super-resolution algorithms. The invertible neural network-based SRFlow [6] outperformed the GAN-based SR algorithms in 2020. Furthermore, the transformer [67] is the dominant network for natural language processing, image classification, and detection, which facilitates researchers to implement the transformer in super-resolution [7]. Additionally, the denoising diffusion model outperforms the GAN in various perceptual metrics within the generative computer vision field [68]. The first denoising diffusion model-based SR algorithm was introduced in 2021 [69]. However, these early diffusion-based SR algorithms [69; 70] initially faced challenges with slow sampling speeds and required many inference steps. Recently, researchers [2; 3; 55] have successfully developed diffusion-based super-resolution methods that can operate in a single step. Autoregressive models and neural operator-based SR algorithms [8; 9; 71] have also been successfully employed in the SR domain. Our paper introduces a novel family of perceptual loss objectives that improve several state-of-the-art SR algorithms [2; 55; 56] across different metrics.

## 2.2 Perceptual Objectives in Super-Resolution

In the super-resolution literature, various perceptual losses have been proposed to improve realistic texture and edge generation. Initial works utilized a pretrained VGG network [36], alongside multiple training strategies [11] and the inclusion of adversarial loss [5]. Wavelet domain style transfer [14] has improved the perception-distortion trade-off in SR by enhancing low-frequency features and transferring style into the wavelet domain. Frequency domain perceptual loss emphasizes several frequency bands of an image to depict its perceptual quality better [20]. Targeted perceptual loss has been applied in SR, utilizing semantic information (object, background, boundary labels) across different image regions to compute perceptual loss and enhance texture and edge quality [35]. Furthermore, Fourier loss introduces adversarial losses directly in Fourier space to enable perception-oriented SR, allowing a smaller network to achieve comparable perceptual quality [25]. Task-driven perceptual (TDP) loss guides SR networks in restoring high-frequency details relevant to specific recognition tasks [72]. The authors [73] demonstrate that contextual loss approximates KL divergence as a statistical comparison tool for a more effective super-resolution strategy. The authors of EnhanceNet [17] argue that traditional SR methods optimize for pixel-wise accuracy (PSNR) but tend to produce blurry images during SR. Consequently, the authors propose combining adversarial training with perceptual loss and a novel texture-matching loss to facilitate the generation of more realistic textures. Perceptual content losses [12] utilize various perceptual loss functions, including discrete cosine transform coefficient loss and differential content loss, in conjunction with adversarial networks for super-resolution. The SSDNet [74] maps RGB and depth features to spherical space for improved feature decomposition, then fuses and refines the information to achieve depth map super-resolution. The Discrete Cosine Transform (DCT)-based perceptual loss emphasizes structural information that is sensitive to the human visual system [75]. FreqNet [21] uses the DCT to learn and reconstruct high-frequency details, the spatial extraction network (SEN), which extracts and transforms spatial features from the low-resolution input image into frequency-domain features, and a frequency reconstruction network (FRN), which reconstructs the high-frequency details. Our MGHF framework prioritizes, preserves, and regularizes multi-granular information, including details, style, content, and regional characteristics, during super-resolution.

In the subsequent section, we will discuss the different components of the MGHF framework: the invertible neural network-based detailed feature extractor, adaptive filter pruning, and reweighting of the detailed features. We will also address our content-style consistency approach that preserves and regularizes content and style information in the INN domain.

# 3 Methodology

In this section, we will discuss an **a**daptive and **w**eighted **d**etail **f**eature **e**xtractor (AWDFE) and content-style consistency. The building block of AWDFE is an invertible neural network-based detail feature extractor (DFE). A brief discussion will be presented here; however, the topic will be discussed in more detail in the supplementary material.

## 3.1 Detail Feature Extractor

We propose a detailed feature extractor (DFE) trained on the ImageNet [40] dataset for preserving texture, fine-grained, and content information between super-resolution and ground-truth images. The building block of our proposed DFE is the invertible neural network that consists of the affine coupling layer [34]. A brief description of the DFE is provided in Algorithm 1. Let $X_{GT}$ and $X_{LR}$ be the ground-truth and corresponding low-resolution image sample caused by down-sampling, blur, and real-world degradation. Any super-resolution method transforms $X_{LR}$ to $X_{SR}$. The DFE is used to extract detailed feature maps by:

$$\begin{aligned} &\mathbf{G} = \mathbf{DFE}(X_{GT}), \quad && \mathbf{S} = \mathbf{DFE}(X_{SR}), \quad && where \\ &\mathbf{G} = \{G_1, G_2, \ldots, G_L\}, \quad && \mathbf{S} = \{S_1, S_2, \ldots, S_L\}, \quad && L \text{ is the number of DFE feature maps.} \end{aligned} \tag{1}$$

The naive multi-granular high-frequency perceptual loss (MGHF-n) is calculated between DFE features of GT and SR images in the following way:

$$\mathcal{L}_{\text{MGHF-n}} = \mathcal{L}_{\text{MSE}}(\mathbf{G}, \mathbf{S}). \tag{2}$$

**Algorithm 1** Pretraining of Detail Feature Extractor

**Require:** Invertible module ($\psi_k$)-based feature extractor, CNN module ($\mathcal{C}_l$), fully connected layers (FC), convolution ($Conv$) for increasing feature maps from 3 to N.
**Require:** ImageNet training set ($\bar{Z}$)

```
while not converged do
    sample z̄ ∼ (Z̄)
    z_1 = Conv(z̄)
    for k = 1, 2, ..., K do
        z_{k+1} = ψ_k(z_k)
    end for
    ŷ_1 = z_K
    for l = 1, 2, 3, ....L do
        ŷ_{l+1} = C_l(ŷ_l)
    end for
    y_score = Soft-max( FC(ŷ_L))
    L = L_{Cross−Entropy}(y_score, y_class)
    Perform a gradient descent step on ∇_{(Conv,ψ,C_l,FC)} L
end while
return The embedding of the K-th invertible module (z_K).
```

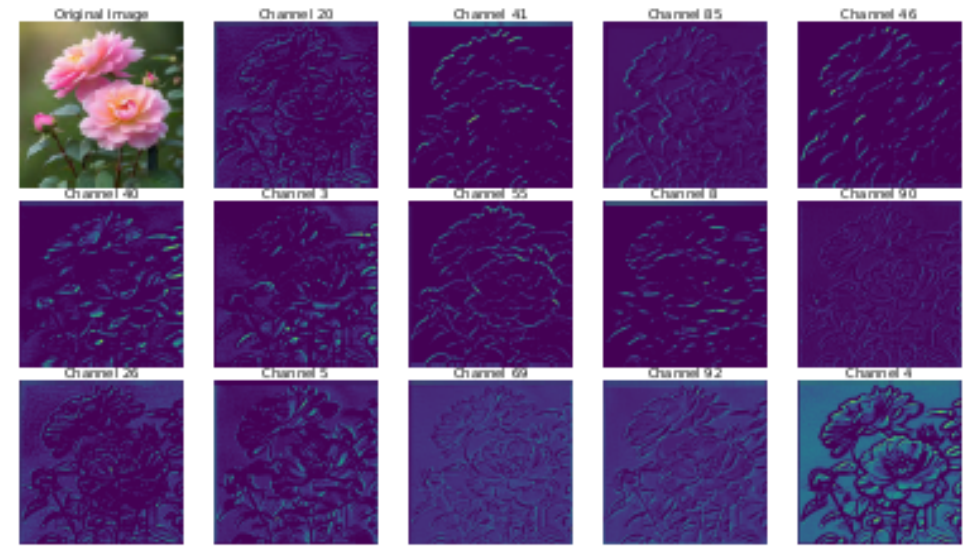

Figure 3: Visualization of detailed feature maps of the pretrained INN-based feature extractor while calculating naive multi-granular high-frequency perceptual loss. (Please zoom in on the figure for better perception.)

### 3.2 Adaptive and Weighted Detail Feature Extractor

The detail feature maps encompass various aspects of an image. However, some of the feature maps consist of less informative and redundant information. To overcome these issues and improve robustness [76; 77] while calculating perceptual loss, we propose adaptive DFE filter weighting and pruning strategies that utilize the entropy calculation on DFE feature maps. The importance score ($I_{\text{combined}}(j)$) of all the extracted DFE feature maps is calculated by using entropy, which enables us to select the most informative M feature maps from the L number of detailed feature maps by using Eq. 4. These M selected feature maps are then weighted by introducing an adaptive weighting strategy in Eq. 5.

$$I_{\text{combined}}(j) = \frac{(1 - H_{\text{norm}}(G_j)) + (1 - H_{\text{norm}}(S_j))}{2}, \tag{3}$$

where $H_{norm}$ is the normalized entropy in the range $[0, 1]$; and $j = 1, 2, ..., L$ feature maps from DFE.

$$\begin{aligned} &\mathcal{M} = \{\text{indices of top } M \text{ feature maps}\} \\ &\hat{\mathbf{G}} = \{G_i \mid i \in \mathcal{M}\} = \{G_{i_1}, \ldots, G_{i_M}\} \text{ , and, similarly, } \hat{\mathbf{S}} \text{ is calculated.} \end{aligned} \tag{4}$$

$$\begin{aligned} w_i &= (1 + \alpha \cdot I_{\text{combined}}(i))^{\gamma}, & i \in \mathcal{M}, & \\ G_i^w &= w_i \cdot G_i, & S_i^w = w_i \cdot S_i, & \quad i \in \mathcal{M}, \\ \mathbf{G}^w &= \{G_{i_1}^w, G_{i_2}^w, \ldots, G_{i_M}^w\}, & \mathbf{S}^w = \{S_{i_1}^w, S_{i_2}^w, \ldots, S_{i_M}^w\}, & \end{aligned} \tag{5}$$

where $\hat{G}$ and $\hat{S}$ are the adaptive ground-truth and super-resolution pruned filters, respectively. $w_i$ is the importance score of $i$-th pruned feature map and $\alpha$ and $\gamma$ are constant. By prioritizing and pruning the detail feature extractor (DFE) outputs based on importance scores, we obtain the adaptive and weighted feature maps $\mathbf{G}^w$ and $\mathbf{S}^w$, constituting our AWDFE module.

### 3.3 Content-Style Consistency

The content-style consistency objective preserves and regularizes the content and style features between ground-truth and super-resolution AWDFE features. While style and content information preservation is widely employed in super-resolution literature [12; 17], we specifically utilize style and content loss in the features of the AWDFE domain by applying the Gram matrix and mean squared error loss. We propose content-style consistency regularization by utilizing correlation loss between SR and GT image pairs in the AWDFE domain.The total **c**ontent-**s**tyle **c**onsistency objective is denoted as ($\mathcal{L}_{\text{CSC}}$) in the following equation:

$$\mathcal{L}_{\text{CSC}} = \beta_1 \cdot \mathcal{L}_{\text{MSE}}(\mathbf{G}^w, \mathbf{S}^w) + \beta_2 \cdot \mathcal{L}_{\text{corr}}(\mathbf{G}^w, \mathbf{S}^w) + \beta_3 \cdot \mathcal{L}_{\text{Gram}}(\mathbf{G}^w, \mathbf{S}^w), \tag{6}$$

where $\mathcal{L}_{\text{corr}}$, $\mathcal{L}_{\text{Gram}}$, and $\mathcal{L}_{\text{MSE}}$ are content-style consistency regularizer, style, and content loss, respectively.

$$\mathcal{L}_{\text{corr}}(\mathbf{G}^w, \mathbf{S}^w) = 1 - \frac{1}{M} \sum_{i=1}^{M} \frac{\text{cov}(G_i^w, S_i^w)}{\sigma_{G_i^w} \cdot \sigma_{S_i^w}}, \quad \mathcal{L}_{\text{Gram}}(\mathbf{G}^w, \mathbf{S}^w) = \frac{1}{M} \sum_{i=1}^{M} \|\text{Gram}(G_i^w) - \text{Gram}(S_i^w)\|^2. \tag{7}$$

### 3.4 Local Information Preservation Objective

Unpaired image-to-image domain translation [78] is a well-known technique in the computer vision literature for transferring modalities; we assume super-resolution and ground truth modalities as two distinct modalities during the training procedure. To transfer GT to SR modality, we utilize the modulated patch-wise noise contrastive estimation (MoNCE) [54] that effectively facilitates regional texture transfer. The proposed local information preservation objective is calculated between DFE feature maps of SR and GT modalities, which can be depicted as:

$$\begin{aligned}\mathcal{L}_{LIP} &= \frac{1}{L}\sum_{k=1}^{L}\mathcal{L}_{MoNCE}(G_k, S_k),\\ &= \frac{1}{L}\sum_{k=1}^{L}\left\{-\sum_{i=1}^{N_k}\log\left[\frac{e^{(s_{ki}\cdot g_{ki}/\tau)}}{e^{(s_{ki}\cdot g_{ki}/\tau)} + Q(N_k-1)\sum_{\substack{j=1\\ j\neq i}}^{N_k} a_{ij}^{k} e^{(s_{ki}\cdot g_{kj}/\tau)}}\right]\right\},\end{aligned} \tag{8}$$

where $L$ is the number of feature maps from DFE, each feature map is divided into $N_k$ patches, and each patch is projected into the embedding space. $a_{ij}$ is the weighting factor for a negative patch that is calculated through the Sinkhorn optimal transport plan [79]. The supplementary material section describes the mathematical equations and parameter settings associated with this objective.

### 3.5 Total Objective

Our proposed MGHF-c framework optimizes the MGHF-n, content-style consistency, and local information preservation objectives. The overall objective can be defined as:

$$\mathcal{L}_{\text{MGHF-c}} = \Gamma_1 \cdot \mathcal{L}_{\text{MGHF-n}} + \Gamma_2 \cdot \mathcal{L}_{\text{CSC}} + \Gamma_3 \cdot \mathcal{L}_{\text{LIP}}, \tag{9}$$

where $\Gamma_1$, $\Gamma_2$ and $\Gamma_3$ are hyperparameters to balance the overall super-resolution process in multifarious granularity.

## 4 Experiment

### 4.1 Experimental Setup

**Training Details.** While fine-tuning different models with the MGHF objective, we adopt the same model architecture and parameter setup as [2; 55; 56; 70]. Specifically, we fine-tune the OSEDiff [2], SinSR [55], and BSRGAN [56] models using the MGHF framework. For all models, we follow the real-world degradation pipeline from RealESRGAN [80]. We maintain the original training protocols and datasets for each model: SinSR is trained on ImageNet [40], while OSEDiff uses the LSDIR [81] dataset combined with the first 10K face images from FFHQ [82]. We fine-tune OSEDiff+MGHF for two epochs and SinSR+MGHF for 20K iterations. For training both BSRGAN and BSRGAN+MGHF from scratch, we use the LSDIR [81] dataset and the first 10K face images from FFHQ [82] for five epochs.

In the total objective equation 9, we determine the optimal values of $\Gamma_1$, $\Gamma_2$, and $\Gamma_3$ to be 2, 1.5, and $10^{-3}$, respectively. Our experiment is conducted on two workstations, each with two NVIDIA RTX A6000 GPUs.

**Training Detail Feature Extractor.** We train our detail feature extractor (based on an invertible neural network [83]) alongside convolutional and fully-connected layers to calculate MGHF perceptual loss. Initially, we use a convolutional block [84] to expand the image feature map from 3 to $N(=128)$. The N channel of an image then passes through an invertible neural network. We take the output from the detail feature extractor to calculate MGHF-n perceptual loss. This network is trained on the ImageNet [40] dataset. We train this model for 20 epochs with a learning rate of 5e-4 with a batch size of 32 and an exponential scheduler with a factor of 0.95 every 5000 iterations. This model is optimized by Adam [85] optimizer.

**Compared methods.** We analyze the performance of our proposed method with several super-resolution algorithms, e.g., StableSR-s200 [86], RealSR-JPEG [87], DiffBIR-s50 [88], SeeSR-s50 [89], OSEDiff [2], PASD-s20 [90], ESRGAN [65], ResShift [70], SinSR [55], BSRGAN [56], SwinIR [91], RealESRGAN [80], DASR [92], and LDM [93].

**Metrics.** We employ PSNR, SSIM, DISTS [94], and LPIPS [11] metrics for performance analysis on the testing dataset with reference images. Fréchet Inception Distance (FID) [95] measures the distribution distance between ground-truth and generated images. Furthermore, we utilize four widely used non-reference image quality metrics to evaluate SR images' realism and semantic coherence: CLIPIQA [51], MUSIQ [52], MANIQA [96], and NIQE [53].

| Methods | Datasets | | | |
|---|---|---|---|---|
| | *RealSR*† | | *RealSet65* | |
| | CLIPIQA↑ | MUSIQ↑ | CLIPIQA↑ | MUSIQ↑ |
| ESRGAN [65] | 0.2362 | 29.048 | 0.3739 | 42.369 |
| RealSR-JPEG [87] | 0.3615 | 36.076 | 0.5282 | 50.539 |
| BSRGAN [56] | 0.5439 | **63.586** | 0.6163 | **65.582** |
| SwinIR [91] | 0.4654 | 59.636 | 0.5782 | 63.822 |
| RealESRGAN [38] | 0.4898 | 59.678 | 0.5995 | 63.220 |
| DASR [92] | 0.3629 | 45.825 | 0.4965 | 55.708 |
| LDM-15 [93] | 0.3836 | 49.317 | 0.4274 | 47.488 |
| ResShift-15 [70] | 0.5958 | 59.873 | 0.6537 | 61.330 |
| *SinSR-1* [55] | **0.6887** | 61.582 | **0.7150** | 62.169 |
| *SinSR-1* + ***MGHF-n*** | **0.7240** | **61.897** | **0.7405** | **63.966** |

† RealSR is preprocess similar procedure as [55] .

Table 1: Quantitative comparison among different super-resolution models on two real-world datasets. The best and the second best results among the SR methods are highlighted in **red** and **blue** colors, respectively.

| Methods | Metrics | | | | |
|---|---|---|---|---|---|
| | PSNR↑ | SSIM↑ | LPIPS↓ | CLIPIQA↑ | MUSIQ↑ |
| ESRGAN [65] | 20.67 | 0.448 | 0.485 | 0.451 | 43.615 |
| RealSR-JPEG [87] | 23.11 | 0.591 | 0.326 | 0.537 | 46.981 |
| BSRGAN [56] | 24.42 | 0.659 | 0.259 | 0.581 | **54.697** |
| SwinIR [91] | 23.99 | 0.667 | 0.238 | 0.564 | 53.790 |
| RealESRGAN [80] | 24.04 | 0.665 | 0.254 | 0.523 | 52.538 |
| DASR [92] | 24.75 | **0.675** | 0.250 | 0.536 | 48.337 |
| LDM-30 [93] | 24.49 | 0.651 | 0.248 | 0.572 | 50.895 |
| LDM-15 [93] | **24.89** | 0.670 | 0.269 | 0.512 | 46.419 |
| ResShift-s15 [70] | **24.90** | **0.673** | 0.228 | 0.603 | 53.897 |
| SinSR-s1 [55] | 24.56 | 0.657 | **0.221** | **0.611** | 53.357 |
| *SinSR-1* +***MGHF-n*** | 24.31 | 0.645 | **0.225** | **0.660** | **55.323** |

Table 2: Quantitative comparison among widely used super-resolution models on *ImageNet-Test*. The best and second best results are highlighted in **red** and **blue**, respectively.

### 4.2 Experimental Results and Comparison with State-of-the-Art

Table 3: Quantitative comparison with state-of-the-art methods on both synthetic and real-world benchmarks. 's' denotes the number of diffusion reverse steps in the method. The best and second best results of each metric are highlighted in **red** and **blue**, respectively. Highlighted skyblue, lightgreen, and orange color rows are different variants of the SR algorithm with our MGHF framework.

| Datasets | Methods | PSNR↑ | SSIM↑ | LPIPS↓ | DISTS↓ | FID↓ | NIQE↓ | MUSIQ↑ | MANIQA↑ | CLIPIQA↑ |
|---|---|---|---|---|---|---|---|---|---|---|
| DIV2K-Val | StableSR-s200 | 23.26 | 0.5726 | 0.3113 | 0.2048 | **24.44** | 4.7581 | 65.92 | 0.6192 | 0.6771 |
| | DiffBIR-s50 | 23.64 | 0.5647 | 0.3524 | 0.2128 | 30.72 | 4.7042 | 65.81 | 0.6210 | 0.6704 |
| | SeeSR-s50 | 23.68 | 0.6043 | 0.3194 | **0.1968** | **25.90** | 4.8102 | **68.67** | **0.6240** | **0.6936** |
| | PASD-s20 | 23.14 | 0.5505 | 0.3571 | 0.2207 | 29.20 | **4.3617** | **68.95** | **0.6483** | **0.6788** |
| | ResShift-s15 | **24.65** | **0.6181** | 0.3349 | 0.2213 | 36.11 | 6.8212 | 61.09 | 0.5454 | 0.6071 |
| | BSRGAN* | 22.67 | 0.5717 | 0.4428 | 0.2839 | 90.74 | 4.6398 | 58.92 | 0.4231 | 0.6268 |
| | BSRGAN*+MGHF-c | 23.27 | 0.5922 | 0.3910 | 0.2569 | 69.16 | **3.9963** | 62.54 | 0.4949 | 0.5875 |
| | SinSR-s1 | **24.41** | 0.6018 | 0.3240 | 0.2066 | 35.57 | 6.0159 | 62.82 | 0.5386 | 0.6471 |
| | SinSR +MGHF-c | 24.25 | 0.6100 | 0.3393 | 0.2202 | 50.78 | 5.6939 | 62.53 | 0.5208 | 0.6708 |
| | OSEDiff-s1 | 23.72 | 0.6108 | **0.2941** | 0.1976 | 26.32 | 4.7097 | 67.97 | 0.6148 | 0.6683 |
| | OSEDiff +MGHF-c | 24.17 | **0.6227** | **0.2813** | **0.1934** | **25.44** | 4.6774 | 67.60 | 0.6221 | 0.6506 |
| DrealSR | StableSR-s200 | 28.03 | 0.7536 | 0.3284 | 0.2269 | 148.98 | 6.5239 | 58.51 | 0.5601 | 0.6356 |
| | DiffBIR-s50 | 26.71 | 0.6571 | 0.4557 | 0.2748 | 166.79 | 6.3124 | 61.07 | 0.5930 | 0.6395 |
| | SeeSR-s50 | 28.17 | 0.7691 | 0.3189 | 0.2315 | 147.39 | 6.3967 | **64.93** | 0.6042 | 0.6804 |
| | PASD-s20 | 27.36 | 0.7073 | 0.3760 | 0.2531 | 156.13 | **5.5474** | **64.87** | **0.6169** | 0.6808 |
| | ResShift-s15 | **28.46** | 0.7673 | 0.4006 | 0.2656 | 172.26 | 8.1249 | 50.60 | 0.4586 | 0.5342 |
| | BSRGAN* | 26.79 | 0.7580 | 0.4027 | 0.2839 | 224.89 | **5.9202** | 53.18 | 0.4334 | 0.6067 |
| | BSRGAN*+MGHF-c | 27.66 | 0.7895 | 0.3454 | 0.2497 | 198.54 | 5.9792 | 58.20 | 0.4956 | 0.5552 |
| | SinSR-s1 | 28.36 | 0.7515 | 0.3665 | 0.2485 | 170.57 | 6.9907 | 55.33 | 0.4884 | 0.6383 |
| | SinSR +MGHF-c | 28.10 | 0.7759 | 0.3334 | 0.2488 | 185.78 | 6.8817 | 57.51 | 0.4967 | **0.6813** |
| | OSEDiff-s1 | 27.92 | **0.7835** | **0.2968** | **0.2165** | **135.30** | 6.4902 | 64.65 | 0.5899 | **0.6963** |
| | OSEDiff +MGHF-c | **28.70** | **0.7942** | **0.2821** | **0.2129** | **128.49** | 6.6821 | 64.54 | **0.6051** | 0.6788 |
| RealSR | StableSR-s200 | 24.70 | 0.7085 | 0.3018 | 0.2288 | 128.51 | 5.9122 | 65.78 | 0.6221 | 0.6178 |
| | DiffBIR-s50 | 24.75 | 0.6567 | 0.3636 | 0.2312 | 128.99 | 5.5346 | 64.98 | 0.6246 | 0.6463 |
| | SeeSR-s50 | 25.18 | 0.7216 | 0.3009 | 0.2223 | 125.55 | 5.4081 | **69.77** | **0.6442** | 0.6612 |
| | PASD-s20 | 25.21 | 0.6798 | 0.3380 | 0.2260 | 124.29 | 5.4137 | 68.75 | **0.6487** | 0.6620 |
| | ResShift-s15 | **26.31** | **0.7421** | 0.3460 | 0.2498 | 141.71 | 7.2635 | 58.43 | 0.5285 | 0.5444 |
| | BSRGAN* | 24.02 | 0.6830 | 0.3949 | 0.2716 | 218.79 | **5.1710** | 59.67 | 0.4424 | 0.6350 |
| | BSRGAN*+MGHF-c | 24.95 | 0.7207 | 0.3416 | 0.2463 | 185.26 | **5.2761** | 64.49 | 0.5314 | 0.5572 |
| | SinSR-s1 | **26.28** | 0.7347 | 0.3188 | 0.2353 | 135.93 | 6.2872 | 60.80 | 0.5385 | 0.6122 |
| | SinSR +MGHF-c | 25.82 | 0.7397 | 0.3069 | 0.2419 | 148.88 | 5.9970 | 62.94 | 0.5430 | **0.6792** |
| | OSEDiff-s1 | 25.15 | 0.7341 | **0.2921** | **0.2128** | **123.49** | 5.6476 | **69.09** | 0.6326 | **0.6693** |
| | OSEDiff +MGHF-c | 26.01 | **0.7418** | **0.2731** | **0.2057** | **111.54** | 5.6058 | 68.32 | 0.6419 | 0.6571 |

**Quantitative comparisons on real-world datasets.** We evaluate the performance of our proposed MGHF framework on three real-world datasets: RealSR[38], RealSet65 [70], and DrealSR [37]. We investigate the image perceptual quality of MGHF compared with other state-of-the-art super-resolution algorithms in Table 1and 3. As shown in Table 1, by applying our MGHF-n to SinSR, we achieve the best CLIPIQA [51] score among widely used GAN, transformer, and diffusion model-based SR algorithms on RealSR and RealSet65 datasets. We also analyze various reference and non-reference metrics of diffusion model-based approaches compared to ours on the DrealSR[37] and RealSR[2] datasets in Table 3. In the RealSR dataset, OSEDiff+MGHF-c achieves the best LPIPS, DISTS, and FID scores, while SinSR+MGHF-c achieves the best CLIPIQA score compared to others. Furthermore, in the DrealSR dataset, OSEDiff+MGHF-c achieves the best PSNR, SSIM, LPIPS, DISTS, and FID scores compared to other approaches. However, for some non-reference image quality metrics such as CLIPIQA and MUSIQ, the performance improvement from MGHF-c shows a slight diminution compared to the performance improvement observed in other metrics (PSNR, RealSR is preprocessed similar procedure as [2].[†] VGG-based Perceptual Loss [57].[††] LPIPS loss [11].

Table 4: Ablation study of proposed methods on synthetic and real-world three benchmark datasets. The best result is depicted in bold format. (MGHF-n, CSC, and LIP represent naive multi-granular high-frequency perceptual loss, content-style consistency, and local information preservation objectives, respectively.)

| Datasets | Methods | NIQE↓ | MUSIQ↑ | CLIPIQA↑ |
|---|---|---|---|---|
| DIV2K-Val | SinSR-s1 [55] | 6.02 | 62.82 | 0.6471 |
| | SinSR-1 +Perceptual Loss[†] | 5.97 | 61.94 | 0.6713 |
| | SinSR-1+LPIPS[††] | 6.06 | 62.95 | 0.6638 |
| | *SinSR-1 +**MGHF-n*** | **5.80** | **63.69** | **0.6822** |
| RealSet65 | SinSR-s1 [55] | 5.98 | 62.17 | 0.7150 |
| | SinSR-1 +Perceptual Loss[†] | 5.63 | 62.64 | 0.7343 |
| | SinSR-1+LPIPS[††] | 5.84 | 63.70 | 0.7295 |
| | *SinSR-1 +**MGHF-n*** | **5.54** | **63.97** | **0.7405** |
| RealSR[*] | SinSR-s1 [55] | 6.29 | 60.80 | 0.6122 |
| | SinSR-1 +Perceptual Loss[†] | 6.15 | 62.43 | 0.6670 |
| | SinSR-1 + LPIPS[††] | 6.36 | 61.84 | 0.6580 |
| | *SinSR-1 +**MGHF-n*** | 6.02 | 62.85 | 0.6740 |
| | *SinSR-1 +**MGHF-n+CSC*** | 6.01 | 62.89 | 0.6731 |
| | *SinSR-1 +**MGHF-n+CSC+LIP*** | **5.99** | **62.94** | **0.6792** |

SSIM, LPIPS, DISTS, and MANIQA). Overall, we can conclude that MGHF-c consistently improves performance across numerous metrics when applied to the OSEDiff, SinSR, and BSRGAN baseline models.

**Quantitative comparisons on synthetic datasets.** We investigate the reference-based fidelity metrics and non-reference-based image quality metrics in the ImageNet-Test [40] and DIV2K-Val [97] datasets. From Table 2, the SinSR+MGHF-n method achieves the best MUSIQ and CLIPIQA scores and the second-best LPIPS score compared to the nine other SR approaches in the ImageNet-Test dataset, though SinSR+MGHF-n lags slightly in PSNR and SSIM metrics. We also find on the DIV2K-val dataset from Table 3 that MGHF-c significantly improves performance on numerous metrics, e.g., SSIM, LPIPS, DISTS, FID, when applied to the OSEDiff, SinSR, and BSRGAN baseline models.

**Qualitative comparisons.** We visually compare two samples with and without the use of MGHF on OSEDiff [2], SinSR [55], and BSRGAN [56] in Figure 4. In the first sample, we can see that the MGHF framework improves the bird's crown with more realistic details compared to the OSEDiff [2] baseline, as the proposed invertible neural network trained on ImageNet provides an important prior for SR. Similarly, in the second sample, our MGHF framework increases detailed information on the window of the castle compared to the OSEDiff baseline. We also found that MGHF consistently improves the detailed information in both samples when applied to the SinSR and BSRGAN models.

### 4.3 Ablation Study

**Effectiveness of naive multi-granular high-frequency (MGHF-n) perceptual loss.** The effectiveness of the proposed MGHF-n perceptual loss can be deduced from the quantitative comparison in Tables 1, 2, and 4. All these results depict the efficacy of MGHF-n in the SinSR algorithm. Specifically, CLIPIQA [51] is significantly improved by the naive MGHF objective.

**Effectiveness of content-style consistency (CSC) and local information preservation (LIP) objective in MGHF.** We systematically add the content-style consistency (CSC) and local information preservation (LIP) objectives to the MGHF-n framework while training on SinSR [55]. The effect of these objectives is depicted in Table 4. From this table, we can conclude that both CSC and LIP modules improve the super-resolution performance in NIQE, MUSIQ, and CLIPIQA metrics on the RealSR dataset. This experiment demonstrates the importance of content-style consistency and local information preservation objectives in the MGHF-c framework.

**Comparison of MGHF with LPIPS and naive perceptual loss.** We compare the efficacy of the proposed MGHF-n and MGHF-c with VGG-based naive perceptual loss [10] and LPIPS [11] on DIV2K-Val, RealSet65, and RealSR test sets. From Table 4, we can deduce that simple MGHF-n outperforms both VGG-based naive perceptual loss and LPIPS on these datasets when implemented in SinSR [55]. This comparison is performed using NIQE, MUSIQ, and CLIPIQA metrics across two real-world datasets and one synthetic dataset. Furthermore, MGHF-c outperforms MGHF-n on the RealSR dataset.

**Parameter and Computational Complexity of Detail Feature Extractor.** Our INN-based detailed feature extractor (DFE) provides substantial efficiency improvements compared to the conventional VGG16-based perceptual loss [10; 11] model. While the standard VGG16-based perceptual loss requires $56.13$ MB of memory with $14,714,688$ parameters, consuming $7.64$ MegaFLOPs per inference and throughput of $215.22$ inferences per second, our DFE architecture achieves comparable improvement with only $1.31$ MB of memory, $343,616$ parameters, and $0.34$ MegaFLOPs, while improving throughput to $400.91$ inferences per second.

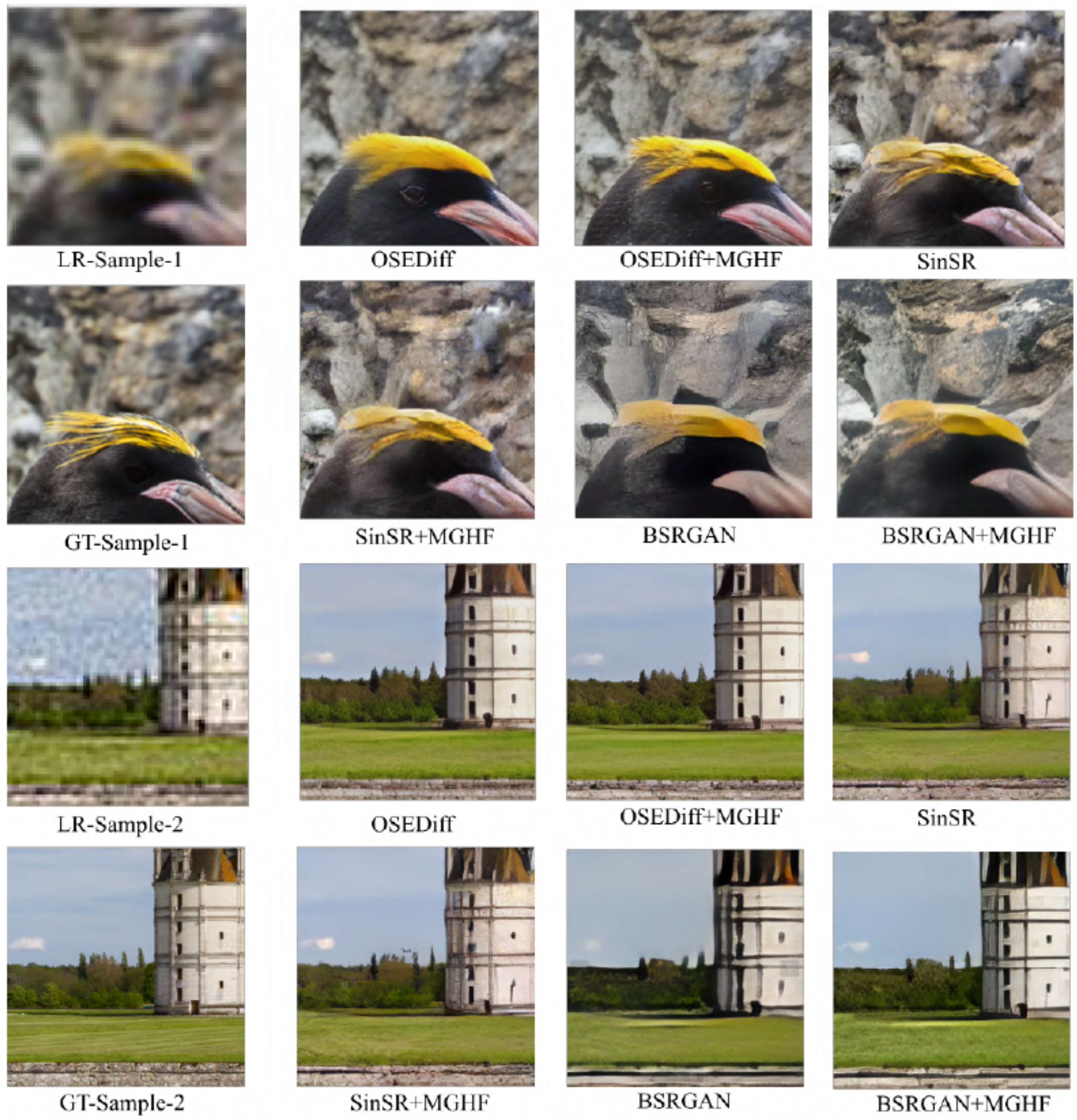

Figure 4: Qualitative comparisons of different SR methods with and without MGHF framework. Please zoom in for detailed information depiction.

## 5 Conclusion and Limitation

We introduce a novel multi-granular high-frequency perceptual (MGHF) framework with a family of losses to preserve multifaceted detail, content, and local features during image super-resolution. For this purpose, we propose an invertible neural network (INN) feature extractor, which significantly improves perceptual loss calculations compared to traditional CNN methods. Leveraging our INN-based super-resolution prior, we adaptively prioritize important features through an entropy-based importance score and preserve content-style consistency through correlation loss. Additionally, we reformulate the super-resolution problem as an unpaired image-to-image translation task by utilizing modulated PatchNCE, which efficiently transfers local information from the ground-truth to the super-resolution modality. Both qualitative and quantitative results demonstrate the efficacy of our proposed MGHF framework when implemented with GAN and diffusion-based super-resolution models.

Our MGHF framework demonstrated promising performance gains on OSEDiff, SinSR, and BSR-GAN. In the future, we will apply the MGHF framework to FLOW, autoregressive, transformer, and neural operator-based super-resolution models. One limitation of this research is that the local information preservation loss is a computationally intensive module due to the modulated PatchNCE objective. Therefore, we implemented it only in the final fine-tuning step.

# A Appendix

In the appendix, we provide the following materials:

- Elaboration of the invertible neural network-based detail feature extractor (referring to Section 3.1 in the main paper).
- Preliminary discussion of local information preservation objective, therefore we discussed PatchNCE and Modulated PatchNCE (referring to Section 3.4 in the main paper).
- Additional visual comparisons of real-world and synthetic samples are shown under a $4 \times 4$ scaling factor (see Section 4.2 in the main paper).

## A.1 Detail Feature Extractor

We utilize an invertible neural network (INN) to capture high-frequency detail features in our proposed MGHF framework. In the NICE paper [98], researchers first proposed the INN concept. The authors of RealNVP [34] subsequently developed the *affine coupling layer*, which enabled more efficient and straightforward data inversion. Utilizing 1×1 invertible convolution, the Glow paper [99] demonstrated generation of realistic high-resolution images. INNs have been applied beyond generation—they've improved classification tasks through superior feature extraction capabilities and information-preserving properties [100]. Moreover, the INN-based detail feature extractor is also used in visible-infrared image fusion [83] and sensor fusion [101] literature. Let $X_{GT}$ and $X_{LR}$ be the ground-truth and corresponding low-resolution image sample caused by down-sampling, blur, and real-world degradation. Any super-resolution method transforms $X_{LR}$ to $X_{SR}$. The DFE is used to extract detailed feature maps by:

$$\begin{aligned} &\mathbf{G} = \mathbf{DFE}(X_{GT}), \qquad \mathbf{S} = \mathbf{DFE}(X_{SR}), \qquad where \\ &\mathbf{G} = \{G_1, G_2, \dots, G_L\}, \quad \mathbf{S} = \{S_1, S_2, \dots, S_L\}, \quad L \text{ is the number of DFE feature maps.} \end{aligned} \tag{10}$$

where $G$ and $S$ represent detail features extracted from the ground-truth and super-resolution images, respectively. The invertible module in the DFE consists of affine coupling layers [34]. The illustration of the invertible module is in Figure 5. In this figure, $\psi^S_{I,l}\,[1\!:\!c]$ is the first $c$ channels of the input feature at the $l$-th invertible layer, where $l = 1, \cdots, L$. The arbitrary mapping functions in each invertible layer are: $\mathcal{I}_1$, $\mathcal{I}_2$, and $\mathcal{I}_3$. We utilize the shallow CNN [102] module as an arbitrary mapping function in the invertible module. Moreover, $G = \psi_{I,L}(X_{GT})$. Finally, the extraction of $S = \mathbf{DFE}(X_{SR}) = \psi_{I,L}(X_{SR})$ can be calculated in the same way as $G$.

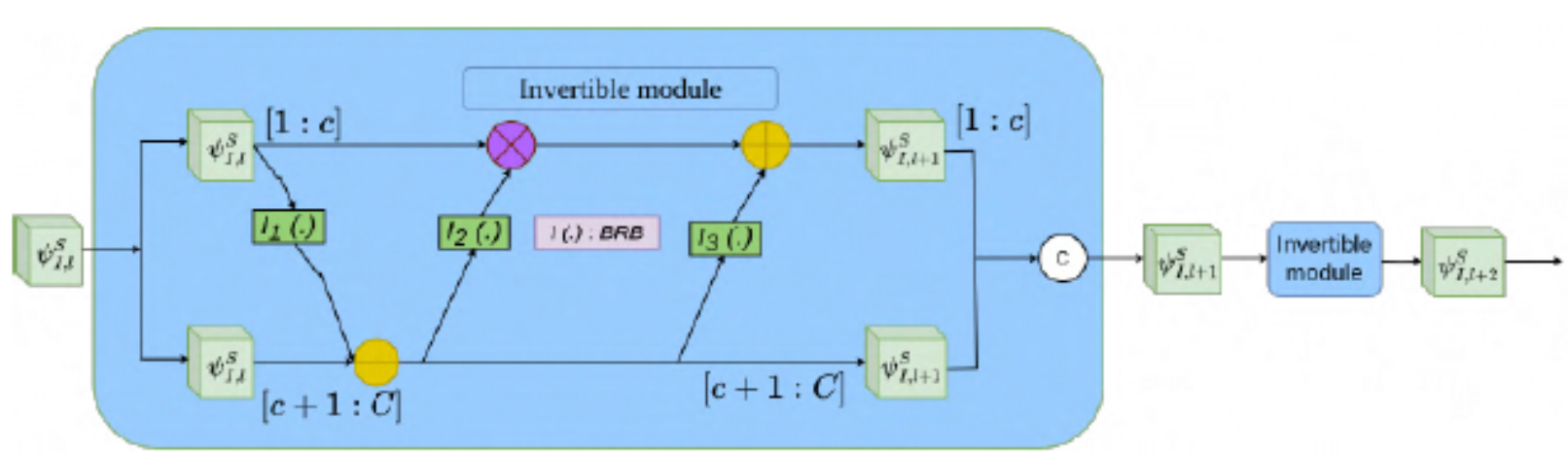

Figure 5: The architecture of the invertible module in the detail feature extractor (DFE) when calculating the multi-granular high-frequency perceptual framework. The DFE consists of $L$ cascaded invertible modules [34]. Each invertible module has an affine coupling layer consisting of scaling and translation functions and a $\odot$ Hadamard product. We use shallow CNN blocks to conduct the scaling and translation operations. Each invertible module contains three shallow CNN blocks.

## A.2 PatchNCE Objective

We introduce a local information preserving (LIP) objective in our MGHF framework. The building block of MGHF is the modulated PatchNCE objective. To understand this, we will first discuss the naive PatchNCE objective. The CUT [103] was one of the pioneering works that introduced a method to maximize the mutual information between the input patch and the corresponding output patch to preserve the semantic content in an unpaired I2I translation [78] scheme by utilizing a contrastive learning framework. After that, several research studies [54; 104; 105] have improved the CUT architecture. The PatchNCE objective maximizes the mutual information, $I(X, Y) = H(X) - H(X|Y)$, which is equivalent to minimizing the conditional entropy $H(X|Y)$. The PachNCE objective can be denoted as:

$$\mathcal{L}_{Patch-NCE}(X, \bar{Y}) = -\sum_{i=1}^{N} log[\frac{e^{(\bar{y}_i . x_i/\tau)}}{e^{(\bar{y}_i . x_i/\tau)} + \sum_{\substack{j=1 \\ j \neq i}}^{N} e^{(\bar{y}_i . x_j/\tau)}}], \tag{11}$$

where $\tau$ is a temperature parameter, and $\bar{Y}$ and $X$ are the generated target domain and ground truth images, respectively. $X = [x_1, x_2, \ldots, x_N]$ and $\bar{Y} = [\bar{y}_1, \bar{y}_2, \ldots, \bar{y}_N]$ represent encoded feature vectors from the 1st, 4th, 8th, 12th, and 16th layers of the encoder. Afterward, these features are passed through a two-layer MLP network [54; 103; 106]. Unlike PatchNCE, we introduce feature maps from every layer of the detail feature extractor while calculating our proposed LIP objective A.4.

In the standard PatchNCE objective, N-class classification is performed where the anchor applies the same contrastive force on all $N - 1$ negative patches, which is often too stringent and detrimental for convergence [54]. To address this issue, we utilize the modulated contrast NCE loss [54] when calculating our local information preservation loss.

### A.3 Modulated Patch-wise Noise Contrastive Estimation Objective

In the contrastive learning literature, the hardness of negative samples has been addressed adequately in [105; 107; 108]. In contrastive learning literature, hard negatives have facilitated the learning of data representations [107]. The hardness of negative patches in unpaired image translation is defined by their similarity to the query [54]. As shown in Eq. 12, hard negative weighting defines the similarity between a negative sample $x_j$ and an anchor $\bar{y}_i$:

$$a_{ij} = \frac{e^{(\bar{y}_i . x_i/\beta)}}{\sum_{j=1}^{N} e^{(\bar{y}_i . x_j/\beta)}}, \tag{12}$$

where $\beta$ is the weighting temperature parameter. The modulated NCE objective employs reweighing procedures by implementing the constraint represented by the following equation:

$$\sum_{i=1}^{N} a_{ij} = 1, \sum_{j=1}^{N} a_{ij} = 1; i, j \in [1, N]. \tag{13}$$

Considering the optimal transport [109], Eq. 14 provides the primary framework, subject to the constraints of Eq. 13.

$$\min_{a_{ij}, i, j \in [1,N]} [\sum_{i=1}^{N} \sum_{\substack{j=1 \\ j \neq i}}^{N} a_{ij}.e^{\bar{y}_i . x_j/\tau}], \tag{14}$$

$$\min_{T} \langle C, T \rangle \;\; s.t \;\; \langle T\overrightarrow{1} \rangle = 1, \; \langle T^T\overrightarrow{1} \rangle = 1, \tag{15}$$

where $\langle C, T \rangle$ is the inner product of the cost matrix ($C$) and transport plan ($T$). In the unpaired I2I network and local information preservation objective, the cost matrix is $e^{\bar{y}_i . x_j/\beta}$ where $i \neq j$; if $i = j$ then $C_{ij} = \infty$. The Sinkhorn [79] algorithm is applied to Eq. 15 for calculating the optimal transport plan. Furthermore, while calculating the modulated contrastive objective in our LIP loss, we use every layer of feature maps of the detail feature extractor. The examples of vanilla and modulated contrast are depicted in Figure 6(a). and Figure 6(b). The MoNCE objective($\mathcal{L}_{MoNCE}$) can be expressed as:

$$\mathcal{L}_{MoNCE} = -\sum_{i=1}^{N} log[\frac{e^{(\bar{y}_i . x_i/\tau)}}{e^{(\bar{y}_i . x_i/\tau)} + Q(N-1)\sum_{\substack{j=1 \\ j \neq i}}^{N} a_{ij} e^{\bar{y}_i . x_j/\tau}}], \tag{16}$$

where $Q$ denotes the weight of negative terms in the denominator and typically $Q = 1$.

## A.4 Local Information Preservation Objective

We assume super-resolution and ground truth modalities are two distinct modalities during the training. To transfer GT to SR modality, we utilize the modulated patch-wise noise contrastive estimation (MoNCE) [54] that effectively facilitates regional texture transfer. The proposed local information preservation objective is calculated between the detail feature extractor (DFE) feature maps of SR and GT modalities, which can be depicted as:

$$\begin{aligned}\mathcal{L}_{LIP} &= \frac{1}{L}\sum_{k=1}^{L}\mathcal{L}_{MoNCE}(G_k, S_k), \\ &= \frac{1}{L}\sum_{k=1}^{L}\left\{-\sum_{i=1}^{N_k}\log\left[\frac{e^{(s_{ki}\cdot g_{ki}/\tau)}}{e^{(s_{ki}\cdot g_{ki}/\tau)} + Q(N_k-1)\sum_{\substack{j=1\\ j\neq i}}^{N_k} a_{ij}^{k} e^{(s_{ki}\cdot g_{kj}/\tau)}}\right]\right\},\end{aligned} \tag{17}$$

where $L$ is the number of feature maps from DFE, each feature map is divided into $N_k$ patches, and each patch is projected into the embedding space. $a_{ij}$ is the weighting factor for a negative patch that is calculated through the Sinkhorn optimal transport plan [79]. The mathematical framework of MoNCE [54] is elaborately described in A.3.

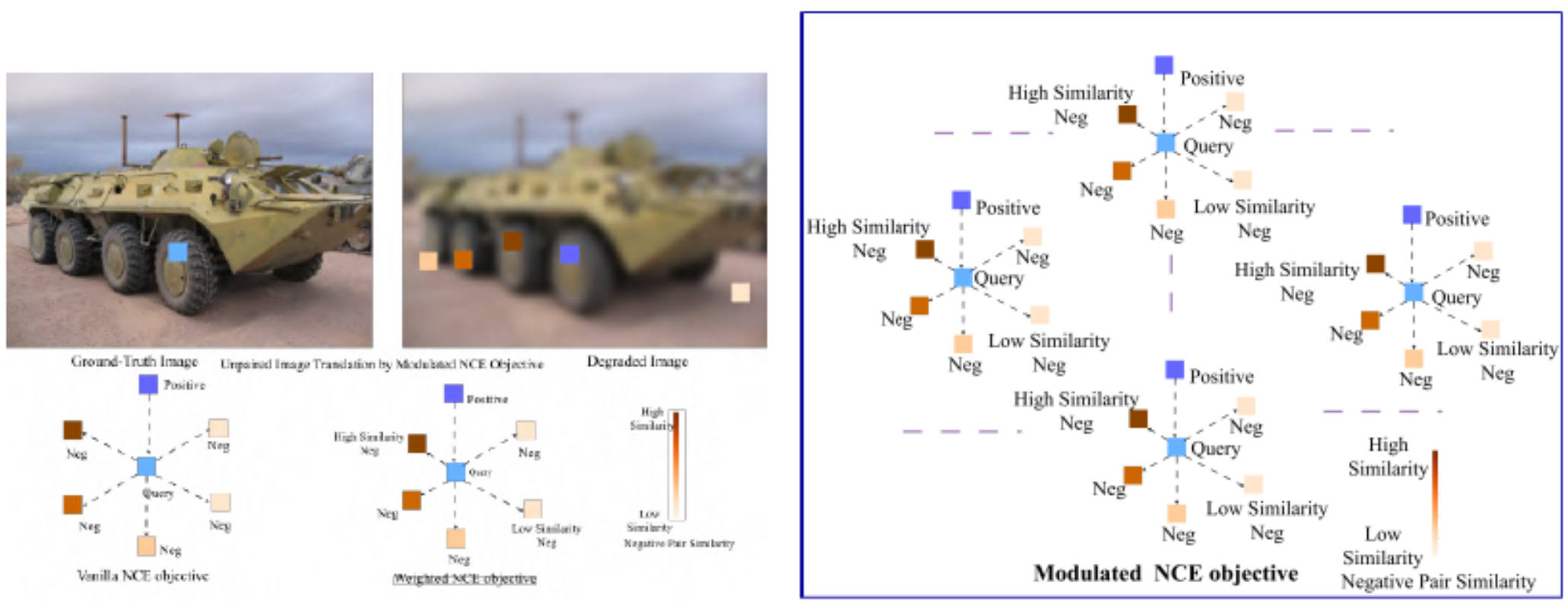

(a) Vanilla and weighted contrastive objective [54]. (b) Modulated contrastive objective [54].

Figure 6: The depiction of modulated contrastive objective [54; 104], which is utilized in our proposed local information preservation objective for image super-resolution.

## A.5 More Parameter Details

In the detail feature extractor, before sending the image to the invertible neural network, we expand the image channels from 3 to $N$. In our experiment, we set $N = 128$. Moreover, in our experiment, we set the number of invertible blocks in the detail feature extractor to one. Furthermore, for the content-style consistency objective, we set $\beta_1 = \beta_3 = 0.1333$ and $\beta_2 = 1$. Finally, in the local information preservation objective, while calculating MoNCE [54], we use $32 \times 32$ patches with a stride of 16 for selecting the neighboring patches.

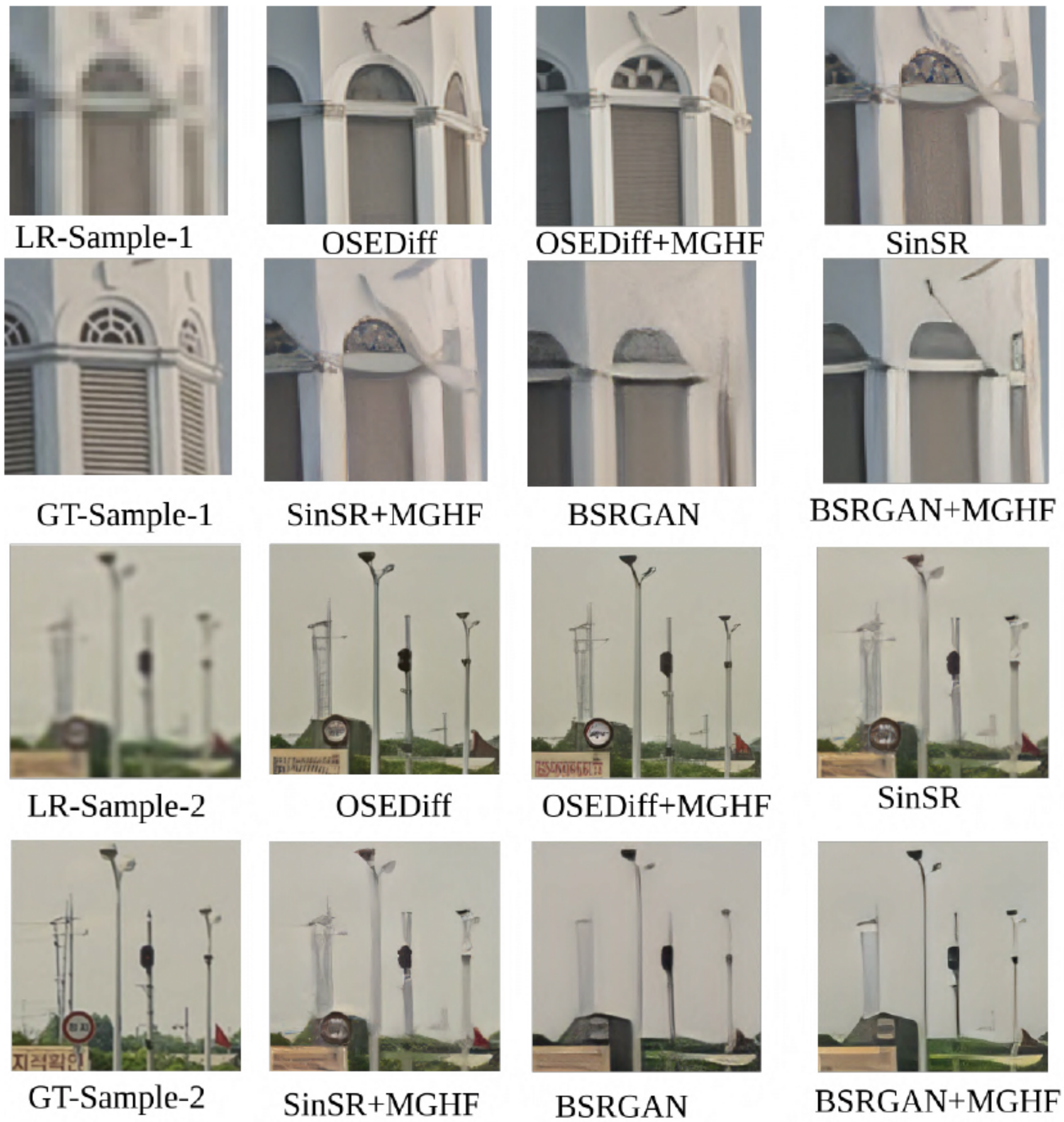

Figure 7: Qualitative comparisons of three state-of-the-art (SOTA) methods with and without the MGHF framework. Zoom in for a clearer view.

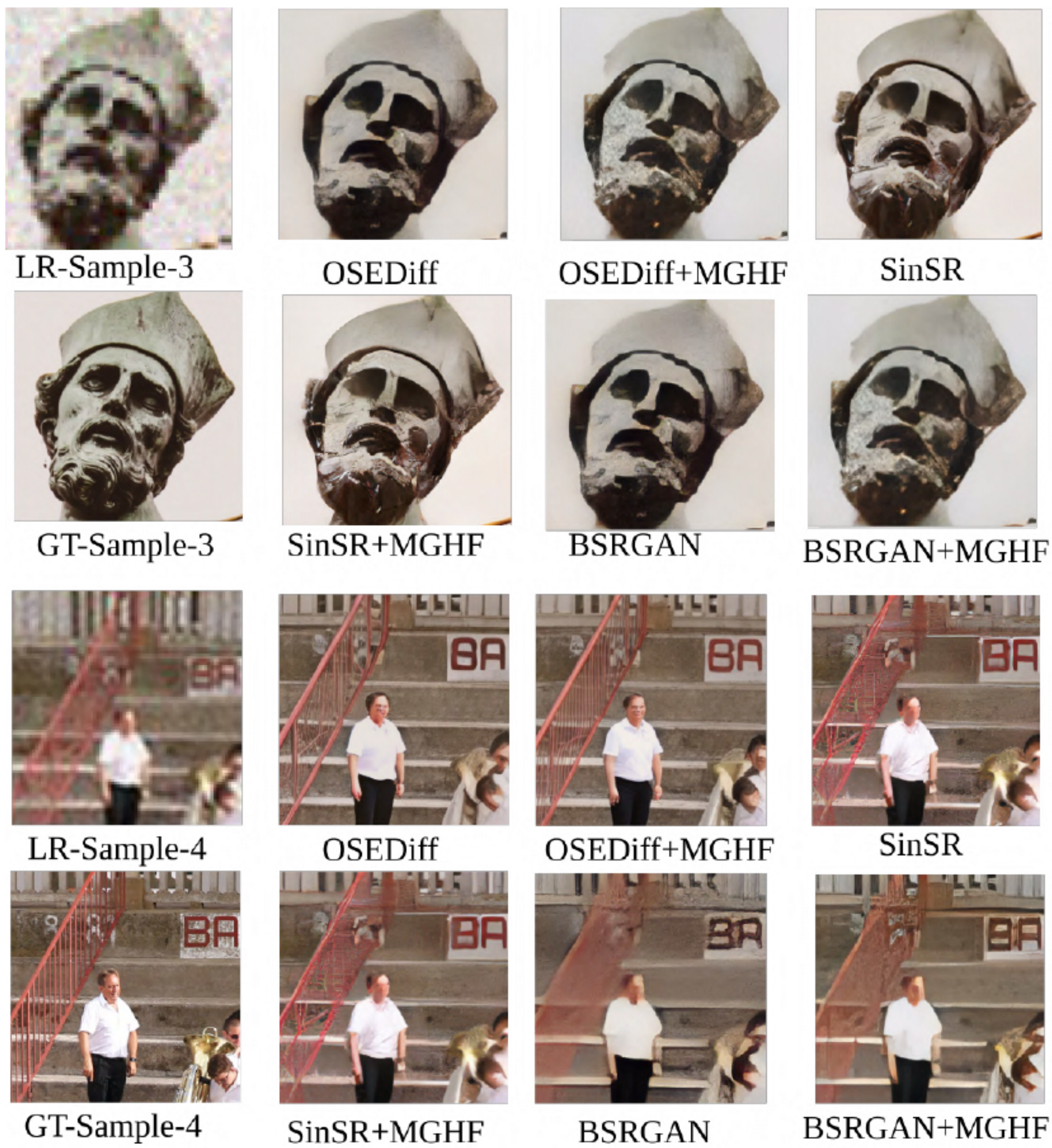

Figure 8: Qualitative comparisons of three state-of-the-art (SOTA) methods with and without the MGHF framework. Zoom in for a clearer view.

## NeurIPS Paper Checklist

1. **Claims**

   Question: Do the main claims made in the abstract and introduction accurately reflect the paper's contributions and scope?

   Answer: [Yes]

   Justification: Yes, the main claims in the abstract and introduction accurately reflect the paper's contributions and scope.

   Guidelines:

   - The answer NA means that the abstract and introduction do not include the claims made in the paper.
   - The abstract and/or introduction should clearly state the claims made, including the contributions made in the paper and important assumptions and limitations. A No or NA answer to this question will not be perceived well by the reviewers.
   - The claims made should match theoretical and experimental results, and reflect how much the results can be expected to generalize to other settings.
   - It is fine to include aspirational goals as motivation as long as it is clear that these goals are not attained by the paper.

2. **Limitations**

   Question: Does the paper discuss the limitations of the work performed by the authors?

   Answer: [Yes]

   Justification: We discuss limitations in the final section (Conclusion and Limitation).

   Guidelines:

   - The answer NA means that the paper has no limitation while the answer No means that the paper has limitations, but those are not discussed in the paper.
   - The authors are encouraged to create a separate "Limitations" section in their paper.
   - The paper should point out any strong assumptions and how robust the results are to violations of these assumptions (e.g., independence assumptions, noiseless settings, model well-specification, asymptotic approximations only holding locally). The authors should reflect on how these assumptions might be violated in practice and what the implications would be.
   - The authors should reflect on the scope of the claims made, e.g., if the approach was only tested on a few datasets or with a few runs. In general, empirical results often depend on implicit assumptions, which should be articulated.
   - The authors should reflect on the factors that influence the performance of the approach. For example, a facial recognition algorithm may perform poorly when image resolution is low or images are taken in low lighting. Or a speech-to-text system might not be used reliably to provide closed captions for online lectures because it fails to handle technical jargon.
   - The authors should discuss the computational efficiency of the proposed algorithms and how they scale with dataset size.
   - If applicable, the authors should discuss possible limitations of their approach to address problems of privacy and fairness.
   - While the authors might fear that complete honesty about limitations might be used by reviewers as grounds for rejection, a worse outcome might be that reviewers discover limitations that aren't acknowledged in the paper. The authors should use their best judgment and recognize that individual actions in favor of transparency play an important role in developing norms that preserve the integrity of the community. Reviewers will be specifically instructed to not penalize honesty concerning limitations.

3. **Theory Assumptions and Proofs**

   Question: For each theoretical result, does the paper provide the full set of assumptions and a complete (and correct) proof?

   Answer: [N/A]

Justification: Our work mainly involves empirical contributions.

Guidelines:

- The answer NA means that the paper does not include theoretical results.
- All the theorems, formulas, and proofs in the paper should be numbered and cross-referenced.
- All assumptions should be clearly stated or referenced in the statement of any theorems.
- The proofs can either appear in the main paper or the supplemental material, but if they appear in the supplemental material, the authors are encouraged to provide a short proof sketch to provide intuition.
- Inversely, any informal proof provided in the core of the paper should be complemented by formal proofs provided in appendix or supplemental material.
- Theorems and Lemmas that the proof relies upon should be properly referenced.

4. **Experimental Result Reproducibility**

   Question: Does the paper fully disclose all the information needed to reproduce the main experimental results of the paper to the extent that it affects the main claims and/or conclusions of the paper (regardless of whether the code and data are provided or not)?

   Answer: [Yes]

   Justification: Yes, the paper fully discloses all necessary information to reproduce the main experimental results, supporting the main claims and conclusions.

   Guidelines:

   - The answer NA means that the paper does not include experiments.
   - If the paper includes experiments, a No answer to this question will not be perceived well by the reviewers: Making the paper reproducible is important, regardless of whether the code and data are provided or not.
   - If the contribution is a dataset and/or model, the authors should describe the steps taken to make their results reproducible or verifiable.
   - Depending on the contribution, reproducibility can be accomplished in various ways. For example, if the contribution is a novel architecture, describing the architecture fully might suffice, or if the contribution is a specific model and empirical evaluation, it may be necessary to either make it possible for others to replicate the model with the same dataset, or provide access to the model. In general. releasing code and data is often one good way to accomplish this, but reproducibility can also be provided via detailed instructions for how to replicate the results, access to a hosted model (e.g., in the case of a large language model), releasing of a model checkpoint, or other means that are appropriate to the research performed.
   - While NeurIPS does not require releasing code, the conference does require all submissions to provide some reasonable avenue for reproducibility, which may depend on the nature of the contribution. For example
     (a) If the contribution is primarily a new algorithm, the paper should make it clear how to reproduce that algorithm.
     (b) If the contribution is primarily a new model architecture, the paper should describe the architecture clearly and fully.
     (c) If the contribution is a new model (e.g., a large language model), then there should either be a way to access this model for reproducing the results or a way to reproduce the model (e.g., with an open-source dataset or instructions for how to construct the dataset).
     (d) We recognize that reproducibility may be tricky in some cases, in which case authors are welcome to describe the particular way they provide for reproducibility. In the case of closed-source models, it may be that access to the model is limited in some way (e.g., to registered users), but it should be possible for other researchers to have some path to reproducing or verifying the results.

5. **Open access to data and code**

Question: Does the paper provide open access to the data and code, with sufficient instructions to faithfully reproduce the main experimental results, as described in supplemental material?

Answer: [Yes]

Justification: All training and testing data are publicly available. We will release the codes and model if the paper is accepted.

Guidelines:

- The answer NA means that paper does not include experiments requiring code.
- Please see the NeurIPS code and data submission guidelines (https://nips.cc/public/guides/CodeSubmissionPolicy) for more details.
- While we encourage the release of code and data, we understand that this might not be possible, so "No" is an acceptable answer. Papers cannot be rejected simply for not including code, unless this is central to the contribution (e.g., for a new open-source benchmark).
- The instructions should contain the exact command and environment needed to run to reproduce the results. See the NeurIPS code and data submission guidelines (https://nips.cc/public/guides/CodeSubmissionPolicy) for more details.
- The authors should provide instructions on data access and preparation, including how to access the raw data, preprocessed data, intermediate data, and generated data, etc.
- The authors should provide scripts to reproduce all experimental results for the new proposed method and baselines. If only a subset of experiments are reproducible, they should state which ones are omitted from the script and why.
- At submission time, to preserve anonymity, the authors should release anonymized versions (if applicable).
- Providing as much information as possible in supplemental material (appended to the paper) is recommended, but including URLs to data and code is permitted.

6. **Experimental Setting/Details**

Question: Does the paper specify all the training and test details (e.g., data splits, hyperparameters, how they were chosen, type of optimizer, etc.) necessary to understand the results?

Answer: [Yes]

Justification: Yes, the paper provides all necessary training and test details, including data splits, hyperparameters, and optimizer type, to fully understand the results.

Guidelines:

- The answer NA means that the paper does not include experiments.
- The experimental setting should be presented in the core of the paper to a level of detail that is necessary to appreciate the results and make sense of them.
- The full details can be provided either with the code, in appendix, or as supplemental material.

7. **Experiment Statistical Significance**

Question: Does the paper report error bars suitably and correctly defined or other appropriate information about the statistical significance of the experiments?

Answer: [NO]

Justification: We do not report error bars.

Guidelines:

- The answer NA means that the paper does not include experiments.
- The authors should answer "Yes" if the results are accompanied by error bars, confidence intervals, or statistical significance tests, at least for the experiments that support the main claims of the paper.
- The factors of variability that the error bars are capturing should be clearly stated (for example, train/test split, initialization, random drawing of some parameter, or overall run with given experimental conditions).

- The method for calculating the error bars should be explained (closed form formula, call to a library function, bootstrap, etc.)
- The assumptions made should be given (e.g., Normally distributed errors).
- It should be clear whether the error bar is the standard deviation or the standard error of the mean.
- It is OK to report 1-sigma error bars, but one should state it. The authors should preferably report a 2-sigma error bar than state that they have a 96% CI, if the hypothesis of Normality of errors is not verified.
- For asymmetric distributions, the authors should be careful not to show in tables or figures symmetric error bars that would yield results that are out of range (e.g. negative error rates).
- If error bars are reported in tables or plots, The authors should explain in the text how they were calculated and reference the corresponding figures or tables in the text.

8. **Experiments Compute Resources**

   Question: For each experiment, does the paper provide sufficient information on the computer resources (type of compute workers, memory, time of execution) needed to reproduce the experiments?

   Answer: [Yes]

   Justification: We detail the type of compute workers, memory, and execution time in the experiment section.

   Guidelines:

   - The answer NA means that the paper does not include experiments.
   - The paper should indicate the type of compute workers CPU or GPU, internal cluster, or cloud provider, including relevant memory and storage.
   - The paper should provide the amount of compute required for each of the individual experimental runs as well as estimate the total compute.
   - The paper should disclose whether the full research project required more compute than the experiments reported in the paper (e.g., preliminary or failed experiments that didn't make it into the paper).

9. **Code Of Ethics**

   Question: Does the research conducted in the paper conform, in every respect, with the NeurIPS Code of Ethics https://neurips.cc/public/EthicsGuidelines?

   Answer: [Yes]

   Justification: Yes, the research in the paper fully conforms to the NeurIPS Code of Ethics in every respect.

   Guidelines:

   - The answer NA means that the authors have not reviewed the NeurIPS Code of Ethics.
   - If the authors answer No, they should explain the special circumstances that require a deviation from the Code of Ethics.
   - The authors should make sure to preserve anonymity (e.g., if there is a special consideration due to laws or regulations in their jurisdiction).

10. **Broader Impacts**

    Question: Does the paper discuss both potential positive societal impacts and negative societal impacts of the work performed?

    Answer: [Yes]

    Justification: Yes, we analyze the potential social impact of super resolution and perceptual loss in the related work, the final section, and the supplementary material.

    Guidelines:

    - The answer NA means that there is no societal impact of the work performed.
    - If the authors answer NA or No, they should explain why their work has no societal impact or why the paper does not address societal impact.

- Examples of negative societal impacts include potential malicious or unintended uses (e.g., disinformation, generating fake profiles, surveillance), fairness considerations (e.g., deployment of technologies that could make decisions that unfairly impact specific groups), privacy considerations, and security considerations.
- The conference expects that many papers will be foundational research and not tied to particular applications, let alone deployments. However, if there is a direct path to any negative applications, the authors should point it out. For example, it is legitimate to point out that an improvement in the quality of generative models could be used to generate deepfakes for disinformation. On the other hand, it is not needed to point out that a generic algorithm for optimizing neural networks could enable people to train models that generate Deepfakes faster.
- The authors should consider possible harms that could arise when the technology is being used as intended and functioning correctly, harms that could arise when the technology is being used as intended but gives incorrect results, and harms following from (intentional or unintentional) misuse of the technology.
- If there are negative societal impacts, the authors could also discuss possible mitigation strategies (e.g., gated release of models, providing defenses in addition to attacks, mechanisms for monitoring misuse, mechanisms to monitor how a system learns from feedback over time, improving the efficiency and accessibility of ML).

11. **Safeguards**

    Question: Does the paper describe safeguards that have been put in place for responsible release of data or models that have a high risk for misuse (e.g., pretrained language models, image generators, or scraped datasets)?

    Answer: N/A

    Justification: The paper poses no such risks.

    Guidelines:

    - The answer NA means that the paper poses no such risks.
    - Released models that have a high risk for misuse or dual-use should be released with necessary safeguards to allow for controlled use of the model, for example by requiring that users adhere to usage guidelines or restrictions to access the model or implementing safety filters.
    - Datasets that have been scraped from the Internet could pose safety risks. The authors should describe how they avoided releasing unsafe images.
    - We recognize that providing effective safeguards is challenging, and many papers do not require this, but we encourage authors to take this into account and make a best faith effort.

12. **Licenses for existing assets**

    Question: Are the creators or original owners of assets (e.g., code, data, models), used in the paper, properly credited and are the license and terms of use explicitly mentioned and properly respected?

    Answer: [Yes]

    Justification: Yes, we clearly indicate the baseline methods and training/testing data used in the paper. Their licenses permit use within the academic scope.

    Guidelines:

    - The answer NA means that the paper does not use existing assets.
    - The authors should cite the original paper that produced the code package or dataset.
    - The authors should state which version of the asset is used and, if possible, include a URL.
    - The name of the license (e.g., CC-BY 4.0) should be included for each asset.
    - For scraped data from a particular source (e.g., website), the copyright and terms of service of that source should be provided.

- If assets are released, the license, copyright information, and terms of use in the package should be provided. For popular datasets, paperswithcode.com/datasets has curated licenses for some datasets. Their licensing guide can help determine the license of a dataset.
- For existing datasets that are re-packaged, both the original license and the license of the derived asset (if it has changed) should be provided.
- If this information is not available online, the authors are encouraged to reach out to the asset's creators.

13. **New Assets**

    Question: Are new assets introduced in the paper well documented and is the documentation provided alongside the assets?

    Answer: [Yes]

    Justification: : Codes and model will be released if the paper is accepted.

    Guidelines:

    - The answer NA means that the paper does not release new assets.
    - Researchers should communicate the details of the dataset/code/model as part of their submissions via structured templates. This includes details about training, license, limitations, etc.
    - The paper should discuss whether and how consent was obtained from people whose asset is used.
    - At submission time, remember to anonymize your assets (if applicable). You can either create an anonymized URL or include an anonymized zip file.

14. **Crowdsourcing and Research with Human Subjects**

    Question: For crowdsourcing experiments and research with human subjects, does the paper include the full text of instructions given to participants and screenshots, if applicable, as well as details about compensation (if any)?

    Answer: N/A

    Justification: The paper does not involve crowdsourcing nor research with human subjects

    Guidelines:

    - The answer NA means that the paper does not involve crowdsourcing nor research with human subjects.
    - Including this information in the supplemental material is fine, but if the main contribution of the paper involves human subjects, then as much detail as possible should be included in the main paper.
    - According to the NeurIPS Code of Ethics, workers involved in data collection, curation, or other labor should be paid at least the minimum wage in the country of the data collector.

15. **Institutional Review Board (IRB) Approvals or Equivalent for Research with Human Subjects**

    Question: Does the paper describe potential risks incurred by study participants, whether such risks were disclosed to the subjects, and whether Institutional Review Board (IRB) approvals (or an equivalent approval/review based on the requirements of your country or institution) were obtained?

    Answer: N/A

    Justification: The paper does not involve crowdsourcing nor research with human subjects

    Guidelines:

    - The answer NA means that the paper does not involve crowdsourcing nor research with human subjects.
    - Depending on the country in which research is conducted, IRB approval (or equivalent) may be required for any human subjects research. If you obtained IRB approval, you should clearly state this in the paper.

- We recognize that the procedures for this may vary significantly between institutions and locations, and we expect authors to adhere to the NeurIPS Code of Ethics and the guidelines for their institution.
- For initial submissions, do not include any information that would break anonymity (if applicable), such as the institution conducting the review.